\documentclass{article}

\usepackage{microtype}
\usepackage{graphicx}
\usepackage{subcaption}
\usepackage{booktabs} 
\usepackage{multirow}

\usepackage[dvipsnames]{xcolor}

\usepackage{aliascnt}

\usepackage{hyperref}

\usepackage[preprint]{icml2026}

\usepackage[utf8]{inputenc} 
\usepackage[T1]{fontenc}    
\usepackage{hyperref}       
\usepackage{url}            
\usepackage{booktabs}       
\usepackage{amsfonts}       
\usepackage{nicefrac}       
\usepackage{microtype}      
\usepackage{xcolor}         

\usepackage{algorithmic}
\usepackage{algorithm}
\usepackage{hyperref}

\usepackage{lmodern}

\usepackage{amsmath}
\usepackage{amssymb}
\usepackage{mathtools}
\usepackage{amsthm}

\usepackage[capitalize,noabbrev]{cleveref}

\theoremstyle{plain}
\newtheorem{theorem}{Theorem}[section]

\newtheorem{property}[theorem]{Property}

\theoremstyle{definition}
\newtheorem{definition}[theorem]{Definition}

\theoremstyle{plain}

\usepackage{bm}
\usepackage{makecell}
\usepackage{tabu, booktabs}
\usepackage{hyperref}
\usepackage{url}
\usepackage{graphicx}
\usepackage{makecell}
\usepackage{float}
\newcolumntype{"}{@{\hskip\tabcolsep\vrule width 1pt\hskip\tabcolsep}}

\usepackage{amsmath,amsfonts,bm}

\def\eqref#1{equation~\ref{#1}}

\def\1{\bm{1}}

\DeclareMathAlphabet{\mathsfit}{\encodingdefault}{\sfdefault}{m}{sl}
\SetMathAlphabet{\mathsfit}{bold}{\encodingdefault}{\sfdefault}{bx}{n}

\DeclareMathOperator{\diag}{diag}

\usepackage{booktabs}
\usepackage{tabularx,ragged2e,siunitx}
\newcolumntype{Y}[1]{>{\Centering\hspace{0pt}\hsize=#1\hsize}X}

\usepackage{caption}

\usepackage{comment}

\usepackage[inline]{enumitem}

\usepackage{xcolor}

\usepackage{listings}

\definecolor{codegreen}{rgb}{0,0.6,0}
\definecolor{codegray}{rgb}{0.5,0.5,0.5}
\definecolor{codepurple}{rgb}{0.58,0,0.82}
\definecolor{backcolour}{rgb}{0.95,0.95,0.92}

\lstdefinestyle{mystyle}{
    backgroundcolor=\color{backcolour},   
    commentstyle=\color{codegreen},
    keywordstyle=\color{magenta},
    numberstyle=\tiny\color{codegray},
    stringstyle=\color{codepurple},
    basicstyle=\ttfamily\footnotesize,
    breakatwhitespace=false,         
    breaklines=true,                 
    captionpos=b,                    
    keepspaces=true,                 
    numbers=left,                    
    numbersep=5pt,                  
    showspaces=false,                
    showstringspaces=false,
    showtabs=false,                  
    tabsize=2
}

\usepackage{wrapfig}

\usepackage{bbm}

\usepackage{booktabs}

\newcommand{\method}{\textsc{InfinityKAN}}

\newcommand{\methodgraph}{\textsc{Infinity-GKAN}}

\usepackage{tcolorbox}
\tcbuselibrary{theorems}

\begin{document}

\twocolumn[
\icmltitle{Variational Kolmogorov-Arnold Network}
\icmltitlerunning{Variational Kolmogorov-Arnold Network}

\icmlsetsymbol{equal}{*}

\begin{icmlauthorlist}
\icmlauthor{Francesco Alesiani}{equal,nec}
\icmlauthor{Henrik Christiansen}{equal,nec}
\icmlauthor{Federico Errica}{equal,nec}
\end{icmlauthorlist}

\icmlaffiliation{nec}{NEC Laboratories Europe, Heidelberg, Germany}

\icmlcorrespondingauthor{Francesco Alesiani}{francesco.alesiani@neclab.eu}
\icmlcorrespondingauthor{Henrik Christiansen}{henrik.christiansen@neclab.eu}
\icmlcorrespondingauthor{Federico Errica}{federico.errica@neclab.eu}

\icmlkeywords{Kolmogorov-Arnold Networks, Variational Inference, Adaptive Basis Functions, Neural Network Architectures}

\vskip 0.3in
]



\printAffiliationsAndNotice{\icmlEqualContribution} 

\begin{abstract}
Kolmogorov-Arnold Networks (KANs) offer a theoretically grounded alternative to multi-layer perceptrons by representing multivariate functions as compositions of univariate basis functions. However, a critical limitation of KANs is the need to manually specify the number of basis functions per layer--a hyperparameter that directly controls model capacity and substantially impacts performance, yet whose optimal value varies unpredictably across tasks. We present \method{}, a variational inference framework that eliminates this design choice by learning the number of basis functions during training. Our approach models the basis count as a latent variable with a truncated exponential prior, introducing a differentiable weighting function that enables gradient-based optimization. We establish the Lipschitz continuity of the variational objective, ensuring stable training dynamics. Experiments across 18 datasets spanning synthetic, image, tabular, and graph domains demonstrate that \method{} matches or exceeds the performance of KANs while requiring no manual selection of the number of bases for each layer. 
\end{abstract}

\section{Introduction}
Kolmogorov-Arnold Networks (KANs) \cite{liu2024kan} have recently gained attention in the machine learning community as a potential alternative to the widely-used Multi-Layer Perceptrons (MLPs) \cite{hornik1989multilayer}. MLPs have been instrumental in transforming machine learning due to their ability to approximate any continuous function, a capability supported by the universal approximation theorem \cite{hornik1989multilayer}. 
The Kolmogorov-Arnold Theorem (KAT), originally developed to address Hilbert's 13th problem, is a fundamental mathematical result with numerous implications \cite{kolmogorov1961representation}. While the universal approximation theorem suggests that any continuous function can be approximated using an MLP of bounded width, KAT represents any multivariate function exactly using a finite and known number of univariate functions.
KAT's influence extends beyond pure mathematics, finding applications in diverse fields such as fuzzy logic, pattern recognition, and neural networks \cite{laczkovich2021superposition, kreinovichNORMALFORMSFUZZY1996, koppen2002training, kuurkova1992kolmogorov,liu2024kan}. This versatility has contributed to its growing importance in the machine-learning community.
KAT-based results have been applied in several ways, including the development of machine learning models, called Kolmogorov-Arnold Networks (KANs) that stand as a potential alternative to MLPs in solving arbitrary tasks \cite{xu2024kaneffectiveidentifyingtracking, decarlo2024kolmogorovarnoldgraphneuralnetworks}.

However, while the KAT argues for the existence of a univariate functions that represent the target function exactly, \textit{the choice of the number of basis functions that model each univariate function remains an open problem}. It is of no surprise that KANs' effectiveness in addressing complex, high-dimensional problems heavily relies on the choice, construction, and training of appropriate basis functions. Various proposals have been made, such as orthogonal polynomials, spline, sinusoidal, or wavelets, which may depend on the specific problem at hand 
\cite{ss2024chebyshevpolynomialbasedkolmogorovarnoldnetworks,mostajeran2024epickanselastoplasticityinformedkolmogorovarnold,bozorgasl2024wavkanwaveletkolmogorovarnoldnetworks, xu2024fourierkangcffourierkolmogorovarnoldnetwork}. Beyond choosing the family of basis functions, the number of basis functions to use is also unknown \textit{a priori}, and an inappropriate choice can substantially degrade a KAN's representational capacity.

We therefore present \method{} which models the univariate functions using an adaptive and potentially infinite number of bases.
\method{} handles the unbounded number of bases in a way that provides gradient information for it to be updated. The model's design stems from a variational treatment of an intractable maximum likelihood learning problem.


\section{Related Works}
\label{sec:related-works}
Recent work \cite{lai2021kolmogorov} has expanded on KAT foundations, exploring the capabilities of KAN-based models in high-dimensional spaces and their potential to mitigate the curse of dimensionality \cite{poggio2022deep}. 
Various KAN architectures have been proposed: KAN has been combined with Convolutional Neural Networks (CNNs)  \cite{ferdausKANICEKolmogorovArnoldNetworks2024}, or with transformer models \cite{yangKolmogorovArnoldTransformer2024}, leading to improved efficiency in sequence modeling tasks.
Furthermore, EKAN incorporates matrix group equivariance \cite{huEKANEquivariantKolmogorovArnold2024a} into KANs, while GKSN \cite{alesianiGeometricKolmogorovArnoldSuperposition2025} explores the extension to invariant and equivariant functions to model physical and geometrical symmetries. 

KANs have demonstrated their versatility across a wide spectrum of machine learning applications \cite{somvanshiSurveyKolmogorovArnoldNetwork2024}, particularly in scenarios demanding efficient (i.e., small number of parameters) function approximation with a limited parameter budget. Their effectiveness in high-dimensional regression problems, where traditional neural networks often face scalability issues, was notably demonstrated by \citet{kuurkova1992kolmogorov}.

Adaptive architectures have been proposed for MLP models. For example, \cite{fahlman_cascade_1989} extends the network with an additional hidden units as the end of a training phase, while firefly network descent \cite{wu_firefly_2020} grows the width and depth of a neural network during training. In continual learning \cite{yoon_lifelong_2018}, network models are updated based on new tasks, or neurons are duplicated or removed according to heuristics to create more capacity \cite{wu_splitting_2019,mitchell_self_2023}. The unbounded depth network of \cite{nazaret_variational_2022}, recently applied to graphs \cite{errica_adaptive_2025}, and adaptive width neural networks \cite{errica2025adaptivewidthneuralnetworks} also use a variational approach to learn the number of neurons of a residual neural network, but these approaches are not directly applicable, since the output is not additive in KAN models.

\section{Infinite Kolmogorov-Arnold Network}
\label{sec:infinity-kan}
We first recap the definition of a KAN layer before introducing our extension to learn an unbounded number of basis functions.

\subsection{KAN layer and basis functions} 
According to the KAT theorem, a generic continuous $d$-dimensional multivariate function $f(x_1,\dots,x_d): \mathbb R^d \to \mathbb R$ defined over a compact space, is represented as a composition of continuous univariate functions as 
\[
f(x_1,\dots,x_d) = \sum_{q=1}^{2d+1} \phi'_q \left( \sum_{p=1}^{d} \phi_{qp}(x_p) \right)
\]
with $ \bm{x} = (x_1,\dots,x_d) \in [0,1]^d$ and where $\phi_{qp},\phi'_q: \mathbb R \to \mathbb R$ are continuous univariate functions. However, the KAN is composed of $L$ KAN layers, 
where each layer $\ell \in \{1,\dots,L\}$ implements the mapping from $[0,1]^{d_{\ell-1}} \to [0,1]^{d_{\ell}}$, where $d_{\ell-1}$ and $d_{\ell}$ are the layer input and output dimensions, using the univariate functions
$\bm{\phi}^\ell = \{ \phi^\ell_{qp} : \mathbb R \to \mathbb R \}$. 
Each layer computes the hidden variables 
$\bm{x}^\ell = \{ x_q^{\ell} ~|~ x_q^{\ell} = h_q^\ell(x^{\ell-1}_1,\dots,x^{\ell-1}_{d_{\ell-1}}) = \sum_{p=1}^{d_{\ell-1}} \phi^\ell_{qp}(x^{\ell-1}_p),  \forall q \in [d_\ell]=\{1,\dots,d_{\ell}\}\}$, from previous layer outputs $\bm{x}^{\ell-1} = \{x^{\ell-1}_p,  \forall p \in [d_{\ell-1}] \}$, i.e., $\bm{x}^{\ell} = \bm{\phi}^\ell(\bm{x}^{\ell-1})$.
The KAT does not tell us how to find the univariate functions $\bm{\phi}^\ell$, but
it is possible to build a convergent series for any uniformly continuous function $\phi(x)$ as a linear combination of 
other base functions $\varphi^n_k(x)$. Therefore 
\begin{align*}
\phi(x) = \lim_{n \to \infty} \phi^n(x), ~~~ \phi^n(x) = \sum_{k=1}^n \phi_k^n(x) = \sum_{k=1}^n \theta_k^n \varphi^n_k(x)    
\end{align*}

where $\varphi_k^n(x)$ can either be a Heaviside step function 
or a rectified linear unit (ReLU) function \cite{jarrett2009best}, as we show in \cref{th:convergence} and \cref{th:relu-annex}, while $\theta_k^n$ are the parameters of the linear combination.
In the following, we refer to $\varphi_k^n(x)$ as the generative functions of the basis $\phi_k^n(x)$.

Therefore, w.l.o.g.\ we represent each univariate function in a KAN layer $\ell$ as the limit of the linear combination of the basis functions $\varphi_k^n(x)$
\begin{align} \label{eq:representation}   
\phi^\ell_{qp}(x) = \lim_{n \to \infty} \sum_{k=1}^{n} \theta^{\ell n}_{qpk} \varphi^n_k(x)
\end{align}

The intuition behind our contribution is that we 
would like to \textit{learn} using a \textbf{finite} number of basis $n$
for each layer that is powerful enough for the task at hand, and therefore training on the finite set of  parameters $\{ \theta^{\ell n}_{qpk} \}_{k \in [n]}$, where $[n]=\{1,\dots,n\}$ for each layer $\ell$, using the simple and efficient back-propagation mechanism. 

\subsection{Variational Training Objective}
\label{sec:variational}
We consider a regression or a classification problem and the corresponding dataset $\mathcal{D}$ composed of i.i.d.\ samples $(\bm{X},\bm{Y}) = \{ ({x}_i,{y}_i)\}_{i=1}^D$, with ${x}_i \in \mathbb{R}^d$ and ${y}_i \in \mathbb{R}^{d'}$. 
If we build a probabilistic model implementing the distribution 
$p(\bm{Y}|\bm{X})$
the objective corresponds to maximize the dataset log-likelihood
\begin{align} \label{eq:obj}
\mathcal{L}(\mathcal{D}) = \ln p(\bm{Y}|\bm{X}) 
= \sum_{i=1}^D \ln p(y_i|x_i).  
\end{align}

\begin{figure}[t]
\centering
\includegraphics[width=0.5\linewidth]{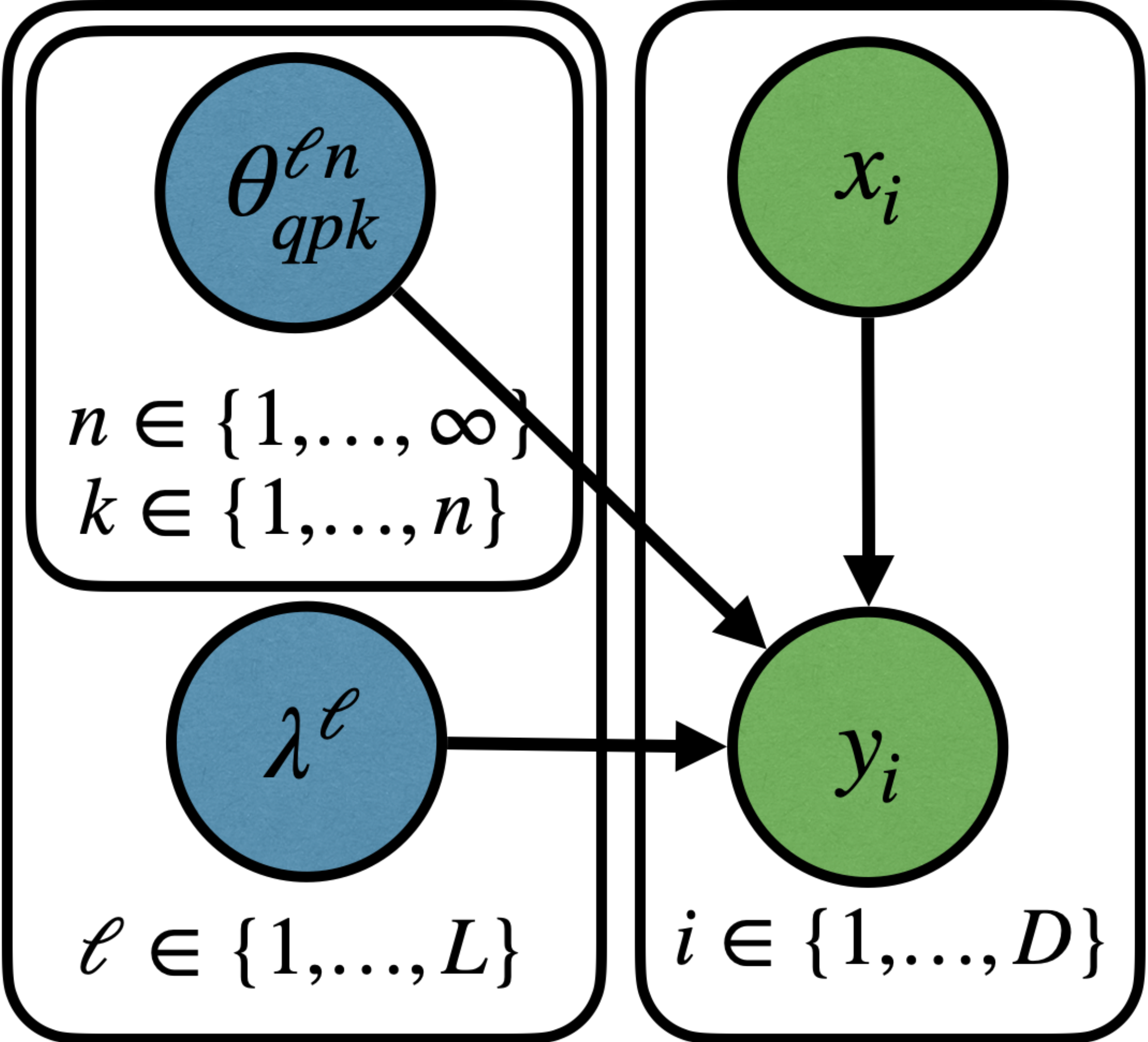}
\caption{
The graphical model of \method{}, with the observable variables (in green) $x_i,y_i$ and latent variables (in blue) $\theta_{qpk}^{\ell n},\nu^\ell$. 
}
\label{fig:infinitykan}
\end{figure}

If we modeled the probability distribution with a multi-layer KAN network, we would need to optimize \cref{eq:obj} with respect to the set of continuous univariate functions 
$\bm{\phi}^\ell = \{\phi^\ell_{qp}\}$. 
However, based on \cref{eq:representation}, we first introduce an infinite-dimensional family of KANs. Because the right value $n$ for each layer is unknown, we introduce two latent variables that parameterize such a family. First, each layer has a set of latent variables 
$\bm{\theta}^\ell= \{ \bm{\theta}^{\ell n} \}_{n=1}^\infty = \{ \theta^{\ell n}_{qpk}, k \in [n] \}_{n=1}^\infty $ 
(see \cref{eq:representation}),
with $\theta^{\ell n}_{qpk}$ is itself a multivariate variable over the learnable weights of the $k$-th basis function at layer $\ell$ and for the $qp$ univariate function. 

We further introduce a latent variable $\nu^\ell$ (the rate parameter of a truncated exponential distribution, see \cite{errica2025adaptivewidthneuralnetworks}) that controls the number of basis functions $n$ used at layer $\ell$. For a KAN of $L$ layers, we define $\bm{\theta}=\left\{ \bm{\theta}^\ell \right\}_{\ell \in [L]}$ 
and $\bm{\nu} = \{ \nu^\ell\}_{\ell \in [L]}$ and we assume independence across all layers, which allows us to write 
$p(\bm{Y}|\bm{X}) = \int d\bm{\theta} d\bm{\nu}p(\bm{Y},\bm{\theta},\bm{\nu}|\bm{X})$.
Similar to \cite{nazaret_variational_2022}, we now assume that $\bm{\theta},\bm{\nu}$ are independent, i.e., $p(\bm{\theta},\bm{\nu})=p(\bm{\theta})p(\bm{\nu})$
and, based on the graphical model of 
\cref{fig:infinitykan}, 
we write the following distributions
\begin{gather}
p(\bm{Y},\bm{\theta},\bm{\nu}|\bm{X})
=
p(\bm{Y}|\bm{\theta},\bm{\nu},\bm{X}) 
p(\bm{\theta})p(\bm{\nu}),\\
p(\bm{\nu}) = \prod_{\ell=1}^L p(\nu_\ell) = \prod_{\ell=1}^L \mathrm{Exp}(\nu_\ell;\eta_\ell), \\
p(\bm{\theta}) 
=
\prod_{
    \substack{
    \ell \in [L],\\
    \mathclap{ n = 1, \dots, \infty, k \in [n], } \\
    \mathclap{q \in [d_{\ell}],p \in [d_{\ell-1}] }
    }    
    } p(\theta^{\ell n}_{qpk}), \\
p(\theta^{\ell n}_{qpk}) 
= 
\mathcal{N}(\theta^{\ell n}_{qpk};\bm{0},\diag{(\sigma_{\ell})})
\end{gather}
where we assume that prior on the rate parameter $\nu$ of the truncated exponential distribution follows an exponential distribution (i.e., $\mathrm{Exp}(\nu;\eta)$), and we further assume that the weights of the basis follow a Gaussian distribution (i.e., $\mathcal{N}(\theta; \mu,\sigma)$). 
The predictive model 
$p(\bm{Y}|\bm{\theta},\bm{\nu},\bm{X})$
is based on the KAN architecture and is described later. 
The distributions depend on the prior's hyper-parameters $\bm{\eta}=\{\eta_\ell\}$ and $ \bm{\sigma}=\{\sigma_\ell\}$, while the KAN is parametrized by $\bm{\theta}$ and $\bm{\nu}$. Maximizing directly \cref{eq:obj} would require computing an intractable integral, therefore, we apply the mean-field variational inference approach \citep{jordan_introduction_1999}, which entails maximizing the expected lower bound (ELBO). By introducing a learnable variational distribution $q(\bm{\theta},\bm{\nu})$ and using the concavity of the logarithmic function, write the objective as (see \cref{sec:ELBO-annex} for the derivation)
\begin{align} \label{eq:elbo}    
     \ln p(\bm{Y}|\bm{X})
     \ge
     \mathbb{E}_{q(\bm{\nu},\bm{\theta})}\left[\ln \frac{p(\bm{Y},\bm{\nu},\bm{\theta} | \bm{X})}{q(\bm{\nu},\bm{\theta})} \right]
\end{align}
Using the same intuition from \cite{nazaret_variational_2022}, we then assume that the variational distribution can be written by conditioning on the number of basis, as
\begin{gather}
    q(\bm{\theta},\bm{\nu}) 
    = 
    q(\bm{\theta}|\bm{\nu})q(\bm{\nu}) \\
    q(\bm{\nu}) 
    =
    \prod_{\ell=1}^L q(\nu_\ell)  = \prod_{\ell=1}^L \mathrm{TruncExp}(\nu_\ell;\bar{\nu}_\ell) \\
    q(\bm{\theta}|\bm{\nu}) 
    =
    \prod_{
    \substack{
    \ell \in [L], \\
    \mathclap{ n  = K_\ell, } \\
    \mathclap{ k \in [K_\ell],} \\
    \mathclap{  q \in [d_{\ell}],p \in [d_{\ell-1}] }
    }
    } q(\theta^{\ell n}_{qpk})
    \prod_{
    \substack{
    \ell \in [L],\\
    \mathclap{ n = 1, \dots, \infty, n \ne K_\ell } \\
    \mathclap{ k \in [n] } \\
    \mathclap{q \in [d_{\ell}],p \in [d_{\ell-1}] }
    }    
    } p(\theta^{\ell n}_{qpk}), \\    
    q(\theta^{\ell n}_{qpk}) 
    =
    \mathcal{N} (\theta^{\ell n}_{qpk}; \bar{\theta}^{\ell n}_{qpk}, \bm{I}), \label{eq:q-theta}
\end{gather}
where $\bar{\nu}_\ell$ and $\bar{\theta}^{\ell n}_{qpk}$ are the variational parameters to be learned, while $K_\ell$ is the truncation number computed from the rate parameter $\nu_\ell$ at a specified quantile threshold $\tau$ (see \cref{sec:weighting-function}), representing the current number of basis functions at layer $\ell$. 

By modeling the distribution of the parameters belonging to a different function in the infinite series with the same \textit{a priori} distribution $p$, its influence on the maximization problem is removed. While we could model the variance of the basis's coefficients with additional trainable parameters, in the following, we see how the variance is ignored. 
We can now write the final objective by using the previous assumptions and the first-order approximation of the expectation, i.e., $
\mathbb{E}_{q(\bm{\nu};\bar{\bm{\nu}})} [f(\bm{\nu})] = f(\bar{\bm{\nu}})
$, and 
$
\mathbb{E}_{q(\bm{\theta}|\bm{\nu};\bar{\bm{\theta}})} [f(\bm{\theta})] = f(\bar{\bm{\theta}})
$, (see \cref{sec:first-oder}) in \cref{eq:elbo}, 
\begin{align} \label{eq:variational-objective}
\sum_{i=1}^D \ln p(y_i | \bm{\nu}=\bar{\bm{\nu}},\bm{\theta}=\bar{\bm{\theta}}, x_i) 
+ \sum_{\ell=1}^L \ln \frac{p(\bar{\nu}_{\ell};\eta_\ell)}{q(\bar{\nu}_{\ell};\bar{\nu}_{\ell})}\\
+ \sum_{
    \substack{
    \ell \in [L], \\
    \mathclap{ k \in [K_\ell],} \\
    \mathclap{  q \in [d_{\ell}],p \in [d_{\ell-1}] }
    }
    } \ln p( \bar{\theta}^{\ell K_\ell}_{qpk} ; \bm{0}, \diag{(\sigma^\ell)}),
\end{align}
where we remove the constant term arising from the evaluation of $q$ distribution at its mean value, i.e., 
$q(\bar{\theta}^{\ell K_\ell}_{qpk}) = \mathcal{N}(\bar{\theta}^{\ell K_\ell}_{qpk}; \bar{\theta}^{\ell K_\ell}_{qpk}, \bm{I}) = \text{const}$ and $\sigma_\ell, \eta_\ell$ are the priors' hyper-parameters. Equipped with \cref{eq:variational-objective}, we can now train the basis parameters $\bm{\bar{\theta}}$ and the rate parameters $\bm{\bar{\nu}}$ using a standard optimization algorithm based on stochastic gradient descent.
The \cref{eq:elbo} contains discrete variables, the number of basis functions. We are therefore faced with two problems: 1) how the gradient propagates, and 2) whether the function is continuous to allow the use of stochastic gradient descent algorithms. We resolve the first question in \cref{sec:weighting-function}, while we provide the following statement for the second, proved in  \cref{sec:new-liptschitz}, 
\begin{tcolorbox}[title=ELBO Lipschitz continuity]
\begin{theorem} 
\label{th:lipschitz}
The ELBO loss of \cref{eq:variational-objective}, with respect to the change in the number of basis $K_\ell$ (or $\nu^\ell$) for the layer $\ell$, is Lipschitz continuous.    
\end{theorem}
\end{tcolorbox}
\subsection{The Weighting Function for the Basis}
\label{sec:weighting-function}
We now introduce the KAN-based model that implements the prediction model 
$p(\bm{Y} | \bm{\nu}=\bar{\bm{\nu}},\bm{\theta}=\bar{\bm{\theta}}, \bm{X})$, given the data samples $\bm{X}$ 
and the variational parameters $\{\bm{\nu},\bm{\theta}\}$. 
The truncation number $K_\ell$ determines how many basis functions to use at layer $\ell$, and is computed from the rate parameter $\nu_\ell$ at a specified quantile threshold $\tau$. To provide gradient information for learning $\nu_\ell$, we introduce a \emph{weighting function} $\bm{w}=\{w_k^{K_\ell}\}_{\ell=1}^L$ parametrized by $\bm{\nu}$ that multiplies the basis coefficients, and write the edge function as
\begin{align} \label{eq:kan-layer}
\phi^\ell_{qp}(x) = 
\sum_{k=1}^{K_\ell}
w_{k}^{K_\ell} \theta^{\ell K_\ell}_{qpk} \varphi^{K_\ell}_k(x)
\end{align}
The weights $w_{k}^{K_\ell}$ are computed as a renormalized probability vector from the truncated exponential distribution:
\begin{align}
    w_k^{K_\ell} &= \frac{P(k)}{\sum_{j=1}^{K_\ell} P(j)}, \quad P(k) = F(k+1) - F(k), \\F(x) &= 1 - e^{-\nu_\ell x}
    \label{eq:weighting-function}
\end{align}
where $F(x)$ is the CDF of the exponential distribution with rate $\nu_\ell$, and $P(k)$ is the discretized probability mass at index $k$. The truncation number is computed as $K_\ell = \lceil -\ln(1-\tau)/\nu_\ell \rceil$ for a specified quantile threshold $\tau$ (e.g., $\tau=0.9$). This formulation ensures that the weights depend differentiably on $\nu_\ell$, allowing gradient-based optimization of the number of basis functions.

\begin{algorithm}[t]
\small
\caption{\method{} Training Procedure}
\begin{algorithmic}[1]
\STATE \textbf{Input:} 
\STATE ~~~~ $\mathcal{D}$: dataset  
\STATE ~~~~ $\mathcal{B}$: basis functions
\STATE \textbf{Output:} Trained \method{} Model $\mathcal{M}$
\STATE Initialize the basis $\mathcal{B}$ and rate parameters $\bm{\nu}$
\FOR{each training epoch}
    \FOR{$(x,y)$ in $\mathcal{D}$}
         \FOR{layer $\ell$ in $\mathcal{M}\text{.KAN\_layers}$}
                \STATE $K_{\ell} \gets \lceil -\ln(1-\tau)/\nu_\ell \rceil$
                \textcolor{gray}{// compute truncation number for truncated exponential}
                \STATE $w_k^{K_\ell} \gets \text{compute\_probability\_vector}(\nu_\ell, K_\ell)$ \textcolor{gray}{// build weights}
                \STATE Update layer parameters if $K_\ell$ changed
        \ENDFOR        
        \STATE $\hat{y} \gets \mathcal{M}(x)$
        \STATE $\text{loss} \gets \text{ELBO}(\mathcal{M}, x, \hat{y})$ \hfill 
        \textcolor{gray}{// \cref{eq:variational-objective} }
        \STATE $\mathcal{M} \gets \text{back-propagation}(\mathcal{M}, \text{loss})$
    \ENDFOR
\ENDFOR

\end{algorithmic}
\label{alg:infinityKANtraining}
\end{algorithm}

\subsection{Updating Weights when the Number of Bases Changes}
\label{sec:interpolation}
When the truncation number $K_\ell$ changes (due to changes in the rate parameter $\nu_\ell$), we update the parameter tensors accordingly. If the number of bases increases from $n$ to $n' > n$, we extend the parameter tensors by initializing new entries with random values (using normal initialization). If the number of bases decreases from $n$ to $n' < n$, we truncate the parameter tensors to keep only the first $n'$ entries. This simple approach avoids the overhead of interpolation while preserving the learned parameters for the overlapping indices.

\section{Experimental validation}
\label{sec:experiments}
\method{} overcomes the limitation of manually selecting the number of basis functions for each layer of a KAN. Our experimental goals are twofold: (1) verify that the training procedure is stable, and (2) confirm that performance is competitive with KANs using a fixed, optimally-tuned number of bases. We focus mostly on classification tasks. In particular, we compare KAN, the propose variational variant (\method{}), and a standard Multi-Layer Perceptrons (MLP) across a broad range of datasets, spanning synthetic, tabular, image, time-series, and graph-like domains.

We consider a total of 18 datasets. Synthetic classification tasks include DoubleMoon, Spiral, and SpiralHard \cite{errica2025adaptivewidthneuralnetworks}, which exhibit increasing levels of nonlinearity and decision-boundary complexity. Image classification benchmarks comprise MNIST, FashionMNIST, CIFAR10, CIFAR100, and EuroSAT \cite{helber2019eurosat}, covering grayscale and color images with varying numbers of classes and degrees of classification complexity. Tabular and time-series datasets include Electricity, House, Jannis, MagicTelescope, MiniBooNE, Phoneme, POL, and EyeMovements \cite{beyazit_inductive_2023}, which vary substantially in dimensionality and sample size. Finally, we include graph-structured datasets NCI1 and REDDIT\_BINARY, where node connectivity and graph topology are essential for classification.

For all tabular datasets, we follow a 70\%/10\%/10\% stratified hold-out split for risk assessment, where the validation set is used for early stopping. We perform model selection by further splitting the training set into 90\% training and 10\% validation sets. We reuse data splits when available. EuroSAT follows the same split percentages as MNIST. Risk assessment results, measured with accuracy, are reported as averages and standard deviations over ten independent runs with different random seeds. For graph datasets, we follow the same experimental setup of \citet{errica_fair_2020} and apply each of the three architectures to a classical GIN baseline \cite{xu_how_2019} to run the experiments.

We report details on the hyper-parameters tried for all models in Section \ref{sec:hyperparameters}.
For the experiments, we used a server with $64$ cores, $1.5$TB of RAM, and $2$ NVIDIA A40 GPUs with $48$GB of RAM.

\section{Results}
In this section, we present the results to evaluate the ability of \method{} to automatically learn the number of bases, and whether the performance of these methods relates to the configuration selected using model selection.
\begin{table}
\centering
\resizebox{\linewidth}{!}{
\begin{tabular}{llll}
\toprule
dataset & KAN & MLP & \method{} \\
\midrule
CIFAR10 & 43.00 (2.05) & 44.03 (1.04) & \textbf{46.99} (0.38) \\
CIFAR100 & 14.04 (0.76) & 15.62 (0.40) & \textbf{19.27} (0.46) \\
DoubleMoon & 98.43 (4.44) & 97.73 (4.57) & \textbf{100.00} (0.00) \\
Electricity & 84.51 (0.40) & 82.05 (0.47) & \textbf{84.54} (0.49) \\
EuroSAT & 68.89 (0.79) & 62.61 (0.59) & \textbf{69.51} (0.32) \\
EyeMovements & \textbf{53.21} (1.15) & 50.80 (0.85) & 50.14 (2.11) \\
FashionMNIST & 85.44 (1.30) & 86.55 (0.20) & \textbf{86.84} (0.30) \\
House & \textbf{89.18} (0.36) & 88.47 (0.43) & 89.11 (0.25) \\
Jannis & \textbf{70.00} (0.28) & 68.15 (0.52) & 68.87 (0.57) \\
MNIST & 96.02 (0.19) & 95.33 (0.24) & \textbf{96.54} (0.20) \\
MagicTelescope & \textbf{88.77} (0.24) & 87.75 (0.39) & 88.18 (0.39) \\
MiniBooNE & 94.12 (0.16) & 94.19 (0.14) & \textbf{94.20} (0.10) \\
NCI1 & 79.30 (1.13) & \textbf{80.00} (1.4) & 76.92 (2.27) \\
POL & 99.20 (0.17) & 99.25 (0.17) & \textbf{99.33} (0.16) \\
Phoneme & \textbf{87.39} (0.77) & 85.99 (1.24) & 87.21 (0.72) \\
REDDIT\_BINARY & \textbf{91.70} (1.17) & 89.9 (1.9) & 83.60 (3.46) \\
Spiral & 99.25 (0.45) & \textbf{99.99} (0.03) & 99.92 (0.07) \\
SpiralHard & \textbf{99.89} (0.06) & 98.88 (1.05) & 98.49 (2.13) \\
\midrule
\textbf{Wins} & 7 & 2 & \textbf{9} \\
\bottomrule
\end{tabular}}
\caption{Comparison of KAN, MLP, and \method{} on 18 datasets. Results are reported as mean accuracy with standard deviation over 10 runs. Bold indicates the best performance for each dataset.}
\label{tab:all-results}
\end{table}

In Table~\ref{tab:all-results} we observe that \method{} outperforms KAN and MLP on 9 out of 18 datasets, with notable improvements on CIFAR10 and CIFAR100. On 7 datasets (EyeMovements, Jannis, MagicTelescope, Phoneme, REDDIT\_BINARY), KAN with a carefully tuned fixed number of bases achieves the best performance, while MLP wins on 2 datasets (NCI1, Spiral, SpiralHard).

Crucially, \method{} achieves this competitive performance \emph{without} requiring the practitioner to tune the number of basis functions--a hyperparameter that KAN requires and that varies from 2 to 128 across datasets in our experiments (see Table~\ref{tab:hyperparameters-bases}). This represents a significant reduction in the hyperparameter search space, as the optimal number of bases is dataset-dependent and not known \textit{a priori}. The results demonstrate that \method{} provides a principled alternative to manual tuning, achieving comparable or superior performance while eliminating an important degree of freedom in model selection.

\begin{table}
\resizebox{\linewidth}{!}{
\begin{tabular}{lcccccc}
\toprule
 & \multicolumn{3}{c}{KAN} & \multicolumn{3}{c}{\method{}} \\
\cmidrule(lr){2-4} \cmidrule(lr){5-7}
dataset & L & d\_emb & Total \# bases & L & d\_emb & Total \# bases \\
\midrule
CIFAR10 & 1 & 16 & 32 & 2 & 16 & 8.00 (0.00) \\
CIFAR100 & 1 & 16 & 2 & 2 & 16 & 23.70 (20.51) \\
DoubleMoon & 1 & 2 & 2 & 1 & 2 & 6.00 (0.00) \\
Electricity & 1 & 16 & 8 & 2 & 16 & 8.00 (0.00) \\
EuroSAT & 2 & 16 & 4 & 2 & 16 & 9.10 (2.02) \\
EyeMovements & 1 & 5  & 128 & 2 & 5 & 33.90 (51.26) \\
FashionMNIST & 2 & 16 & 64  & 2 & 16 & 8.00 (0.00) \\
House & 2 & 5  & 64 & 2 & 5 & 9.30 (0.90) \\
Jannis & 2 & 16 & 16 & 2 & 16 & 8.00 (0.00) \\
MNIST & 2 & 16 & 64 & 2 & 16 & 10.50 (2.20) \\
MagicTelescope & 2 & 5  & 256 & 1 & 16 & 6.00 (0.00) \\
MiniBooNE & 2 & 16 & 16 & 1 & 16 & 6.00 (0.00) \\
NCI1 & 5 & 16 & 10 & 5 & 16 & 14.00 (0.00) \\
POL & 2 & 5  & 4 & 1 & 16 & 21.00 (32.02) \\
Phoneme & 1 & 16 & 128 & 2 & 16 & 23.60 (3.41) \\
REDDIT\_BINARY & 2 & 16 & 16 & 2 & 16 & 88.90 (51.44) \\
Spiral & 1 & 2  & 8 & 1 & 5 & 6.50 (0.67) \\
SpiralHard & 1 & 5  & 8 & 1 & 5 & 9.40 (2.06) \\
\bottomrule
\end{tabular}}
\caption{Comparison of selected hyperparameters and learned total number of bases for KAN and \method{} across datasets. For KAN, L is the number of layers, d\_emb is the embedding dimension, and Total \# bases is the fixed number of bases used. For \method{}, Total \# bases is reported as mean (std) over 10 runs.}
\label{tab:hyperparameters-bases}
\end{table}
Table~\ref{tab:hyperparameters-bases} presents the selected hyperparameters and learned total number of bases for KAN and \method{} across datasets. For KAN, we report the number of layers (L), embedding dimension (d\_emb), and the fixed total number of bases used. For \method{}, we report the learned total number of bases as mean and standard deviation over 10 runs.
We find that \method{} often learns a number of bases that is different from the fixed number used in KAN, demonstrating its ability to adaptively select the appropriate model complexity for each dataset. The variability in the learned number of bases across runs indicates that \method{} can explore different configurations during training, potentially leading to improved performance.
This adaptive behavior highlights the advantage of \method{} in automatically determining the model capacity needed for effective learning, without requiring manual tuning of the number of basis functions.
\par 
\begin{figure*}[ht]
\centering
\includegraphics[width=0.9\linewidth]{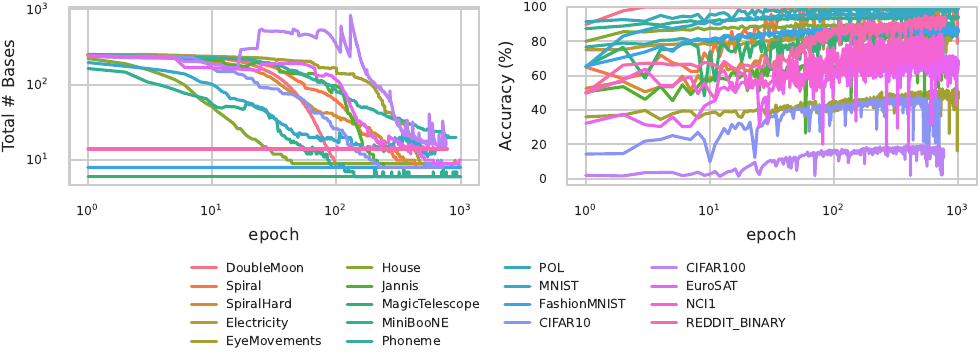}
\caption{Convergence of \method{} during training across all 18 datasets. \textbf{Left:} Evolution of the total number of basis functions (width) over epochs. Most datasets converge to a stable width within the first 100 epochs, demonstrating that \method{} can efficiently identify the appropriate model complexity. \textbf{Right:} Test accuracy (score) convergence over epochs. The adaptive selection of basis functions does not hinder convergence, with most datasets reaching stable performance early in training.}
\label{fig:convergence}
\end{figure*}

For instance, in the CIFAR100 dataset, KAN uses only 2 bases, while \method{} learns an average of 23.7 bases with a high standard deviation, indicating significant variability across runs. In contrast, for the DoubleMoon dataset, both KAN and \method{} use a similar number of bases (2 for KAN and an average of 6 for \method{}), suggesting that a simpler model suffices for this task. 
These examples illustrate how \method{} can flexibly adjust its complexity based on the dataset characteristics, potentially leading to better generalization and performance.
In some instances, such as Electricity and FashionMNIST, \method{} converges to the same number of bases (8) across all runs, indicating a stable selection process for these datasets. This consistency suggests that for certain tasks, there may be an optimal model complexity that \method{} can reliably identify.
However, in other cases like REDDIT\_BINARY, \method{} exhibits high variability in the learned number of bases (mean of 88.9 with a standard deviation of 51.44), reflecting the complexity and diversity of the dataset. This variability may allow \method{} to explore a wider range of model configurations, potentially leading to improved performance on challenging tasks.

Figure~\ref{fig:convergence} illustrates the convergence behavior of \method{} during training across all 18 datasets. The left panel shows the evolution of the total number of basis functions (width) over epochs, while the right panel displays the test accuracy (score) convergence.
We observe that for most datasets, the total number of basis functions stabilizes within the first 100 epochs, indicating that \method{} can efficiently identify the appropriate model complexity early in training. This rapid convergence suggests that the adaptive mechanism effectively guides the model towards a suitable configuration without excessive exploration.
The right panel shows that the test accuracy also converges steadily over epochs, with most datasets reaching stable performance early in training. This indicates that the adaptive selection of basis functions does not hinder convergence, and \method{} can achieve competitive accuracy while dynamically adjusting its complexity.

\begin{figure*}[ht]
\centering
\includegraphics[width=0.9\linewidth]{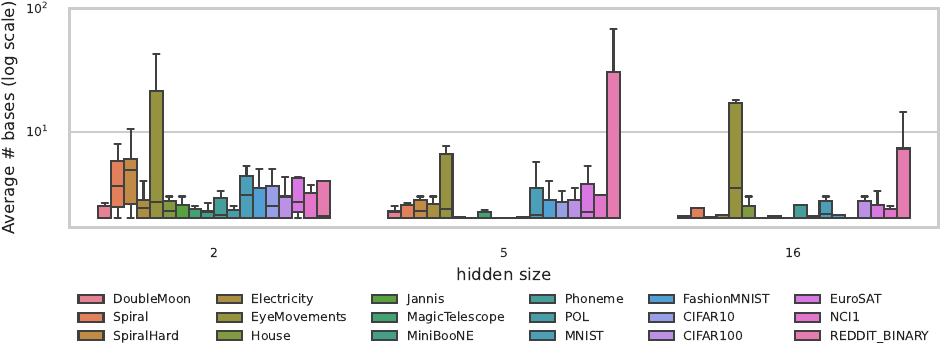}
\caption{Average number of basis functions learned by \method{} across different hidden sizes (2, 5, and 16) for all datasets. The y-axis is shown on a log scale. Error bars indicate standard deviation over 10 runs. The learned number of bases varies across datasets and hidden sizes, with some datasets (e.g., EyeMovements, REDDIT\_BINARY) exhibiting higher variability, while others (e.g., DoubleMoon, Electricity) converge consistently to lower values.}
\label{fig:truncexp_hidden}
\end{figure*}

Figure~\ref{fig:truncexp_hidden} presents the average number of basis functions learned by \method{} across different hidden sizes (2, 5, and 16) for all datasets. The y-axis is shown on a logarithmic scale to better visualize the range of learned bases. Error bars indicate the standard deviation over 10 runs.
We observe that the learned number of bases varies significantly across datasets and hidden sizes. For instance, datasets like EyeMovements and REDDIT\_BINARY exhibit higher variability in the number of learned bases, suggesting that these tasks may require more complex representations and that \method{} explores a wider range of configurations. In contrast, datasets such as DoubleMoon and Electricity consistently converge to lower values of learned bases, indicating that simpler models are sufficient for these tasks.
The results highlight the flexibility of \method{} in adapting its complexity based on the dataset characteristics and the chosen hidden size. This adaptability allows \method{} to effectively balance model capacity and generalization, leading to improved performance across diverse tasks.

\section{Conclusions and future directions}
We have presented \method{}, a variational inference method for training Kolmogorov-Arnold Networks with an adaptive number of basis functions per layer. By formulating basis function selection as a probabilistic inference problem, \method{} automatically determines appropriate model complexity during training, relieving practitioners from manually tuning this critical hyperparameter.

Our theoretical contributions establish the Lipschitz continuity of the ELBO objective with respect to changes in the number of basis functions, ensuring stable optimization dynamics. The proposed weighting function, derived from a truncated exponential distribution, provides differentiable control over the basis selection while maintaining computational efficiency.

Through extensive experiments across 18 diverse datasets--spanning synthetic classification, image recognition, tabular data, time-series, and graph-structured domains--we have demonstrated that \method{} achieves competitive or superior performance compared to KANs with fixed, tuned basis counts and standard MLPs. Notably, \method{} outperforms alternatives on challenging benchmarks such as CIFAR10, CIFAR100, and EuroSAT, while matching performance on simpler tasks like DoubleMoon and Spiral. The convergence analysis reveals that \method{} efficiently identifies appropriate model complexity within the first 100 epochs for most datasets, with stable basis selection across multiple runs.

A key finding is that the learned number of basis functions varies substantially across datasets, reflecting the inherent complexity of different learning tasks. For instance, simpler datasets like Electricity and FashionMNIST consistently converge to 8 basis functions, while more complex datasets like REDDIT\_BINARY and EyeMovements exhibit greater variability, suggesting that these tasks benefit from exploring diverse model configurations.

\paragraph{Outlook and future directions.}
The framework introduced in this work opens several promising avenues for future research. While we have established Lipschitz continuity of the objective, further theoretical analysis could characterize the convergence rates of \method{} and provide formal guarantees on the optimality of the learned basis count, drawing connections to neural architecture search and Bayesian model selection. The current first-order approximation of the variational objective represents a natural starting point, but extending it to more sophisticated inference techniques--such as importance-weighted bounds or normalizing flows--could improve the quality of the learned posterior over basis functions and lead to better uncertainty quantification.

From an architectural perspective, the success of \methodgraph{} on graph-structured data suggests that integrating \method{} with convolutional architectures for image processing, recurrent structures for sequential data, or transformer-based models for language understanding could significantly broaden the applicability of adaptive KANs. Moreover, the learned basis functions and their associated weights offer a unique window into the model's internal representations, which could be leveraged for scientific discovery in domains where understanding functional relationships is as important as achieving predictive accuracy.

Scaling \method{} to very large datasets and models remains an important challenge. Efficient implementations that exploit sparsity patterns in the basis functions and leverage modern hardware acceleration will be crucial for practical deployment in large-scale applications. The adaptive nature of \method{} also makes it a natural candidate for continual learning scenarios, where model complexity can dynamically grow or shrink as new tasks are encountered, and the learned basis structure could serve as a transferable prior for related downstream tasks.

We believe that \method{} represents a significant step toward more principled and automated neural network design, reducing the hyperparameter search space while maintaining or improving predictive performance. By bridging the gap between the theoretical elegance of the Kolmogorov-Arnold representation and practical machine learning, we hope this work will inspire further research into adaptive, interpretable, and efficient neural architectures.

\paragraph{Limitations.}
While \method{} eliminates the need to tune the number of basis functions, it introduces its own hyperparameters (the priors $\lambda$ and $\theta$), though our ablation studies suggest these are less sensitive than the basis count and in the large-data limit become unimportant. The variational optimization introduces computational overhead compared to fixed-basis KANs: in our experiments, \method{} requires approximately $\approx 2\times$ the training time due to the dynamic resizing of parameter tensors. However, note that this overhead is strongly offset by the elimination of hyperparameter search for the number of bases. Additionally, on some datasets (e.g., REDDIT\_BINARY, EyeMovements), \method{} performs comparably to--but does not exceed--carefully tuned KANs, indicating that \method{} can replace a full hyperparameter search but may not outperform a well-guided manual tuning effort.

\section*{Impact Statement}
This paper presents work whose goal is to advance the field of Machine Learning. There are many potential societal consequences of our work, none of which we feel must be specifically highlighted here.



\bibliographystyle{plainnat}
\bibliography{kolmogorov,kan,mlip,scalar,awnn}

\appendix

\section{Theorems, Proofs, and Definitions}
\label{sec:proofs}

\begin{definition} (Uniformly continuous function)
$f$ is uniformly continuous function on $X$, metric space, if  $\forall \epsilon >0$, $\exists \delta >0$ such that $\forall x,y \in X$ and $|x-y| < \delta$,  we have that $|f(x)-f(y)| < \epsilon$.     
\end{definition}

\begin{tcolorbox}[title=Convergence of step and piecewise functions]
\begin{theorem} 
\label{th:convergence}
Let's $f \in C([a,b]=[-1,1], X)$ uniformly continuous on the metric space $X$, and $f_n$ the sequence of step functions, such that
$$
f_n(t) = f(t^{n}_k), ~ t \in [t^n_k,t^n_{k+1}),~ k=1,\dots,n 
$$
or a piece-wise linear function, such that
$$
f_n(t) = f(t^{n}_k)(1-s) + sf(t^{n}_{k+1}),
$$
with $ ~ t \in [t^n_k,t^n_{k+1}),~ k=1,\dots,n$ and $t_1=a=-1 \le t_k \le t_{k+1} \le t_n = 1=b$, with $s = t-t^{n}_k$. Then $f_n$ converges to $f$.
\end{theorem}
\end{tcolorbox}

\begin{tcolorbox}[title=Representation with piecewise linear and Relu functions]
\begin{theorem} 
\label{th:relu-annex}
Any piecewise linear function can be represented as a linear combination of ReLU functions, $g(t)=\max\{0,x\}$, and any uniformly continuous function can be the limit of a sequence of combinations of ReLU functions.  
\end{theorem}
\end{tcolorbox}

\begin{proof}(\cref{th:convergence})
Since $X$ is a metric space, $f_n$ converges uniformly to $f$ iff
$$
\forall \epsilon>0, ~ \exists N \in \mathbb{N}, ~  \forall n \ge N: \|f_n - f\|_\infty < \epsilon
$$
Let's take an $\epsilon >0$ and the corresponding $\delta$ for uniformly continuity of $f$, and choose $N$ such that $(b-a)/N = 2/N \le \delta$, then for $n \ge N$ we have
$$
|f(t_k^n)-f(t)| < \epsilon
$$
and
$$
|f(t)-f(t_{k+1}^n)| < \epsilon
$$
for $t \in [t_k^n,t^n_{k+1})$ and $\| f_n - f\|_\infty \le \epsilon$.
\end{proof}

\begin{proof} (\cref{th:relu-annex})
Following \cref{th:convergence}, we consider the segment $[t^n_k,t^n_{k+1})$, and 
$$
f_n(t) = f(t^{n}_k)(1-s) + sf(t^{n}_{k+1}), ~ t \in [t^n_k,t^n_{k+1}),~ k=1,\dots,n, 
$$
with $s = t-t^{n}_k $, then 
$$
f_n(t) = f(t^{n}_k) + g(t-t_k^n)\frac{f(t^{n}_{k+1})-f(t^{n}_{k})}{\delta_k^n} ~ t \in [t^n_k,t^n_{k+1}), 
$$
with $\delta_k^n = t^n_{k+1}-t^n_k$, and $g(t)=\max \{0,t\}$ the relu function. When we stick together the linear functions, we need to remove the contribution of the previous relu functions in the form of $-\alpha_k^n g(t-t_k^n)$ 
with $\alpha_k^n=-\frac{f(t^{n}_{k})-f(t^{n}_{k-1})}{\delta_{k-1}^n}$. Writing in as a single equation
$$
f_n(t) = f(t_1^n) + \sum_{k=1}^n [g(t-t_k^n)-g(t-t_{k+1}^n)]\frac{[f(t^{n}_{k+1})-f(t^{n}_{k})]}{\delta_k^n} 
$$
\end{proof}

\section{ELBO derivation}
\label{sec:ELBO-annex}
Here, we want to formalize the ELBO derivation.
\begin{tcolorbox}[title=Variational ELBO]
\begin{theorem}
    \label{th:elbo-annex}
Given the assumptions on log likelihood and the variational distribution $q(\bm{\nu},\bm{\theta})$ from \cref{sec:variational}, we have the following ELBO:
\begin{align} \label{eq:elbo-annex}    
     \ln p(\bm{Y}|\bm{X})
     \ge
     \mathbb{E}_{q(\bm{\nu},\bm{\theta})}\left[\ln \frac{p(\bm{Y},\bm{\nu},\bm{\theta} | \bm{X})}{q(\bm{\nu},\bm{\theta})} \right]
\end{align}
\end{theorem}
\end{tcolorbox}

\begin{proof}   
We start from the objective function \cref{eq:obj} and marginalize over the $\bm{\lambda},\bm{\theta}$ variable
\begin{align} 
     \ln p(\bm{Y}|\bm{X})
     = \ln  \int d\bm{\lambda} d\bm{\theta}  p(\bm{Y},\bm{\lambda},\bm{\theta} | \bm{X})
\end{align}
We then divide and multiply by the variational distribution and recognize the expected value against this distribution
\begin{align*} 
     \ln p(\bm{Y}|\bm{X})
     & = \ln  \int d\bm{\lambda} d\bm{\theta}  p(\bm{Y},\bm{\lambda},\bm{\theta} | \bm{X}) \\
     & = \ln  \int d\bm{\lambda} d\bm{\theta} p(\bm{Y},\bm{\lambda},\bm{\theta} | \bm{X}) \frac{q(\bm{\lambda},\bm{\theta} )}{q(\bm{\lambda},\bm{\theta} )} \\
     & = \ln  \int d\bm{\lambda} d\bm{\theta} q(\bm{\lambda},\bm{\theta} )  \frac{p(\bm{Y},\bm{\lambda},\bm{\theta} | \bm{X})}{q(\bm{\lambda},\bm{\theta} )} \\
     & = \ln  \mathbb{E}_{q(\bm{\lambda},\bm{\theta})}
     \left[
     \frac{p(\bm{Y},\bm{\lambda},\bm{\theta} | \bm{X})}{q(\bm{\lambda},\bm{\theta} )} 
     \right] \\     
\end{align*}
We then apply the concavity of the logarithm function 
\begin{align*} 
     \ln  \mathbb{E}_{q(\bm{\lambda},\bm{\theta})}
     \left[
     \frac{p(\bm{Y},\bm{\lambda},\bm{\theta} | \bm{X})}{q(\bm{\lambda},\bm{\theta} )} 
     \right] \\     
     \ge \mathbb{E}_{q(\bm{\lambda},\bm{\theta})}
     \left[
     \ln  \frac{p(\bm{Y},\bm{\lambda},\bm{\theta} | \bm{X})}{q(\bm{\lambda},\bm{\theta} )} 
     \right] \\     
\end{align*}
and we obtain the ELBO
\begin{align} 
     \ln p(\bm{Y}|\bm{X})
     \ge
     \mathbb{E}_{q(\bm{\lambda},\bm{\theta})}\left[\ln \frac{p(\bm{Y},\bm{\lambda},\bm{\theta} | \bm{X})}{q(\bm{\lambda},\bm{\theta})} \right]
\end{align}
\end{proof}

\section{Lipschitz Continuity of the ELBO}
\label{sec:new-liptschitz}
Similar to \cite{errica_adaptive_2025}, we provide a derivation of the Lipschitz continuity of the ELBO. 
Although we provide different approaches on how to smoothly transition in the representation of the univariate function $\bm{\phi}^\ell$ using interpolation, there will still be a sudden change in the KAN functions. 
By providing the following property, we show that this jump is bounded. 
Therefore, we provide a theoretical result (\cref{th:lipschitz}) that shows that the ELBO satisfies the Lipschitz continuity. 

\begin{tcolorbox}[title=ELBO Lipschitz continuity]
\begin{theorem} 
\label{th:lipschitz-annex}
The ELBO loss of \cref{eq:variational-objective}, with respect to the change in the number of basis $K_\ell$ (or $\lambda^\ell$) for the layer $\ell$, is Lipschitz continuous.    
\end{theorem}
\end{tcolorbox}

\begin{proof}    
We focus on the term involving $K_\ell$ of the ELBO, we write \cref{eq:variational-objective} as
$$
\ln p(\bm{Y}|\bar{\bm{\lambda}},\bar{\bm{\theta}}, \bm{X})
+ \ln 
\frac{p(\bar{\bm{\lambda}})}{q(\bar{{\bm\lambda}})}
+ \ln \frac{p(\bar{\bm{\theta}}) }{q(\bar{\bm{\theta}}|\bar{{\bm\lambda}})}
$$
where only the second and last terms depend on $\mathbf K = \{ K_\ell \}_{\ell=1}^L$ . 
Since $K_\ell$ is a deterministic function of $\lambda_\ell$, we consider them, in the following, equivalent. Let's first define
$$
\ln \frac{p(\bar{\bm{\theta}}) }{q(\bar{\bm{\theta}}|\bar{{\bm\lambda}})} = 
\sum_{{\ell}=1}^{L} \sum_{k=1}^{K_{\ell}} 
\ln \frac{p( \bar{ \bm{ \theta } }^{\ell}_k)} {q(\bar{ \bm{ \theta } }^{\ell}_k|{\lambda_\ell})} 
= \sum_{{\ell}=1}^{L} f_1(K_\ell)
$$
We have that $f_1(K_\ell)$ is Lipschitz continuous, indeed, when $K_\ell$ changes to $K'_{\ell}$, we have 
\begin{align}
| f_1(K'_{\ell}) - f_1(K_{\ell})|& 
=| \sum_{k=K_{\ell}}^{K'_{\ell}} 
\ln \frac{p( \bar{ \bm{ \theta } }^{\ell}_k)} {q(\bar{ \bm{ \theta } }^{\ell}_k|{\lambda_\ell})} | \\
& \le 
\sum_{k=K_{\ell}}^{K'_{\ell}} | 
\ln \frac{p( \bar{ \bm{ \theta } }^{\ell}_k)} {q(\bar{ \bm{ \theta } }^{\ell}_k|{\lambda_\ell})} | \\
&\le \max_n | \log \frac{p(\boldsymbol{\rho}_n)}{q(\boldsymbol{\rho}_n|\boldsymbol{\nu})}| |D_{\ell'} - D_{\ell}|
\end{align}
Therefore
$$
| f_1(K'_{\ell}) - f_1(K_{\ell})| \le M |K'_{\ell} - K_{\ell}|
$$
with $M = \max_k | 
\ln \frac{p( \bar{ \bm{ \theta } }^{\ell}_k)} {q(\bar{ \bm{ \theta } }^{\ell}_k|{\lambda_\ell})}
|$. 
We now look at the first term,
$$
f_2(\mathbf K) = \log p(Y | \boldsymbol{\nu},\boldsymbol{\rho}, X)
$$
If we use bounded derivative continuous univariate functions in the KAT representation, and since $f_2$  is the composition of continuous univariate functions, the resulting function is continuous and of bounded derivative and therefore Lipschitz continuous. 
\end{proof}

\section{First-Order approximation and  $n$-th  order approximation}
\label{sec:first-oder}
The first-order approximation requires the function $f \in C^0$ to be continuous in a neighbor of $\mu_x=\mathbb{E}_x[x]$, then
$$
\mathbb{E}_x [f(x)]=\mathbb{E}_x [f(\mu_x)+O((x-\mu_x))] \approx f(\mathbb{E}_x[x])
$$
If $f \in C^1$ we would similarly have  
\begin{align}
\mathbb{E}_x [f(x)]&=\mathbb{E}_x [f(\mu_x)+f'(\mu_x)(x-\mu_x) + O((x-\mu_x)^2)] \\&\approx f(\mathbb{E}_x[x])
\end{align}
while, with $f \in C^2$ we would similarly have
\begin{align}
\mathbb{E}_x [f(x)] & = \mathbb{E}_x [f(\mu_x)+f'(\mu_x)(x-\mu_x) \\&+\frac1{2} f''(\mu_x)(x-\mu_x)^2+O((x-\mu_x)^3)] \\
& \approx f(\mathbb{E}_x[x]) + \frac1{2} f''(\mathbb{E}_x[x])(\mathbb{E}_x[x^2]-\mathbb{E}^2_x[x]) \\&= f(\mu_x) + \frac1{2} f''(\mu_x)\sigma_x^2
\end{align}
The $n$-order approximation
\begin{align}
\mathbb{E}_x [f(x)] & = 
\mathbb{E}_x [
f(\mu_x)
+ \sum_{k=1}^n \frac{f^{(k)}(\mu_x)}{k!}(x-\mu_x)^k 
+ O((x-\mu_x)^{n+1})
] 
\\
& \approx f(\mu_x) 
+ \sum_{k=1}^n \frac{f^{(k)}(\mu_x)}{k!} \mu_x^{(k)} 
\end{align}
with $\mu_x^{(k)}$ the $k$-th momentum of the distribution. 

\section{Lazy Interpolation}
A simpler way to interpolate the weight after a change in the number of basis is to keep the same weights when possible. 
During transition, if we reduce $n \to n' <n$, then we can ignore the parameters $\theta^{\ell n}_k, k=n'+1,\dots,n$ and set $\theta^{\ell n'}_k=\theta^{\ell n}_k, k \in [n']$, while if we increase $n$, then we keep the previous parameters and instantiate the missing ones $\theta^{\ell n'}_k, k=n+1,\dots,n'$.

\section{Chebyshev type-I polynomials}
\label{sec:cheb}
An interesting extension of the framework is when using Chebyshev type-I polynomials as basis functions, indeed the approximation error decreases with the number of bases, and adding bases is probably less critical then with basis on the real line. 
The Chebyshev polynomials of type-I are defined as 
\begin{align} \label{eq:cheb-poly}
T^k(\cos \theta) = \cos (k \theta)
\end{align}
or $T^k(x) = \cos \left( k \cos^{-1} (x) \right), x \in [-1,1]$, where we typically map the real axis to the $[-1,1]$ interval using $z = \tanh(x)$. We can then define the series as 
\begin{align} \label{eq:cheb-act}
    \phi(x) &= \lim_{n \to \infty} \phi^n(x) \\
    \phi^n(x) &= \sum_{k \in [n]} \theta_k T^k(x)
\end{align}
with 
$\theta_k$ 
trainable parameters. 
The interesting point of the use of the Chebyshev polynomial is that we can now share the parameters among series and the dependence on the index $n$ is dropped in the parameters. This is due to the Taylor expansion, where we drop dependence on the higher-order polynomials. 
We then introduce the asymmetric window function
\begin{align}
    w_\lambda(x) &= \left(1 + e^{2(x-\lambda)/\sigma} \right)^{-1} \\
    w_k^n &=  w_\lambda(x_i),~~ x_i \in [\lambda + \sigma]
\end{align}
which then gives the final form of the trainable $\ell$-th KAN layer function
\begin{align}
    \phi_\ell^n(x) &= \sum_{k \in [n]} \theta^{\ell}_k w_k^n T^k(x)
\end{align}
with 
$\bm{\theta}_\ell = \{ \theta^{\ell}_k \}_{k \in [n]}$ 
the trainable parameters, while $\lambda_\ell$ and $\sigma$, the variational parameter and hyperparameter of the variational optimization problem. 
\begin{property}
\label{th:chebyshev}
If we consider $\varphi_k(x) = T^k(x)$ we have that
$$
\int_{-1}^{1} d x h(x) \varphi_k(x) \varphi_{k'}(x) = \delta_{k-k'}
$$
with $h(x) = \frac1{\sqrt{(1-x^2)}}$
\end{property}

\section{Fourier basis and representation}
\label{sec:fourier}
An alternative basis is the one defined based on the Fourier functions
$$
\varphi_k(x) = \frac1{\sqrt T}e^{i  \frac{2\pi}{T} kx}
$$
with $T=2$ and with domain $\Omega = [-1,1]$. We have that any continuous function of period $T$ can be represented as
\begin{align}
\phi(x) &=\lim_{n \to \infty} \sum_{k=-n}^{n} \phi_k(x) \\&=\lim_{n \to \infty} \sum_{k=-n}^{n} \theta_k \varphi_k(x), \\ \phi_k(x) &= \sum_{k=-n}^{n} \theta_k \varphi_k(x) 
\end{align}
with $\theta_k \in \mathbb C$ complex numbers. It is well known that 
\begin{property}(Fourier complex basis )
\label{th:fourier-complex}
If we consider $\varphi_k(x) = \frac1{\sqrt T} e^{i \frac{2\pi}{T} kx}$ we have that
$$
\int_{-T}^{T} d x \varphi_k(x) \varphi^*_{k'}(x) = \delta_{k-k'}
$$
with $\varphi^*_k(x)$ the complex conjugate of $\varphi^*_k(x)$.
\end{property}
If we want to use real numbers, then we have two sets of bases
$$
\varphi_k(x) = \frac1{\sqrt T}\cos{ \left(\frac{2\pi}{T} kx \right)}, ~~~
\varphi'_k(x) = \frac1{\sqrt T}\sin{ \left(\frac{2\pi}{T} kx \right)}
$$
if $T=2$ then
$$
\varphi_k(x) = \frac1{\sqrt 2}\cos{ \left( \pi kx \right)}, ~~~
\varphi'_k(x) = \frac1{\sqrt 2}\sin{ \left(\pi kx \right)}
$$

\begin{property}(Fourier real basis )
\label{th:fourier-real-basis}
If we consider $\varphi_k(x) = \frac1{\sqrt 2}\cos{ \left( \pi kx \right)}, ~~~
\varphi'_k(x) = \frac1{\sqrt 2}\sin{ \left(\pi kx \right)}$ we have that
\begin{align}
\int_{-1}^{1} d x \varphi_k(x) \varphi_{k'}(x) &= \int_{-1}^{1} d x \varphi_k(x) \varphi'_{k'}(x) \\&= \int_{-1}^{1} d x \varphi'_k(x) \varphi'_{k'}(x) = \delta_{k-k'}
\end{align}
\end{property}
When can then use the basis for represent any periodic function in the interval $[-T/2,T/2]$, based on the following property. 
\begin{property}(Fourier representation)
\label{th:fourier-real-representation}
If we consider $\varphi_k(x) = \frac1{\sqrt 2}\cos{ \left( \pi kx \right)}, ~~~
\varphi'_k(x) = \frac1{\sqrt 2}\sin{ \left(\pi kx \right)}$ we have that
$$
\phi(x) = \lim_{n \to \infty } \phi^n(x), ~~~  \phi^n(x) = \sum_{k=0}^{n} \theta_k \varphi_k(x) + \sum_{k=1}^{n} \theta'_k \varphi'_k(x)
$$
with $\theta_k ,\theta'_k $ the coefficients of the series.
\end{property}

\section{Initialization with Chebyshev polynomials}
We first recall that 
\begin{align}
    \int dx T^k(x) = \frac1{2} \left[ \frac{T^{k+1}}{k+1} -\frac{T^{k-1}}{k-1} \right]
\end{align}
therefore
\begin{align}
    \int dx (\phi_\ell^n(x))^2 &= \left(\int dx  \sum_{k \in [n]} \theta^{\ell n}_k w_k^n T^k(x) \right)^2 \\
    &=   \left(\sum_{k \in [n]} \theta^{\ell n}_k w_k^n \int dx T^k(x) \right)^2 \\
    &=   \left(\sum_{k \in [n]} \theta^{\ell n}_k w_k^n \frac1{2} \left[ \frac{T^{k+1}}{k+1} -\frac{T^{k-1}}{k-1} \right] \right)^2 \\
    & \le  \sum_{k \in [n]} (\theta^{\ell n}_k)^2 (w_k^n)^2 \frac1{4} \left( \left[ \frac{T^{k+1}}{k+1} -\frac{T^{k-1}}{k-1} \right] \right)^2 \\
    & \le  \sum_{k \in [n]} (\theta^{\ell n}_k)^2 (w_k^n)^2 \frac1{4}  \\
\end{align}
since $\left| \frac{T^{k+1}}{k+1} -\frac{T^{k-1}}{k-1} \right| \le 1$.
If we want 
$$
\sum_{k \in [n]} (\theta^{\ell n}_k)^2 (w_k^n)^2 \frac1{4} = 1
$$
we can either set the variance to 
$$
\mathbb{E} (\theta_k^n)^2 =  \frac{4}{\sum_{k \in [n]} (w_k^n)^2}  \approx \frac{4}{n-5/4}, 
$$
when we assume the parameters to be i.i.d. and zero mean, $\mathbb{E}[\theta_k^n] = 0$. The $5/4$ term is due to the shape of the windows, when $\sigma=1$, the last sample is $\approx 0$ while the before-last sample is $1/2$.

\newpage
\section{Dataset Characteristics and Hyper-Parameters}
\label{sec:hyperparameters}

Table~\ref{tab:dataset_details} summarizes the dataset properties, including input dimensionality, output dimensionality, and dataset size (number of observations).

\begin{table}[ht]
\setlength{\tabcolsep}{4pt} 
\renewcommand{\arraystretch}{1.1} 
\caption{Dataset characteristics.}
\label{tab:dataset_details}
\centering
\begin{tabular}{lccc}
\toprule
           & Samples & Features & Classes \\ 
           \midrule
CIFAR10    & 60000      & 32x32x3     & 10  \\
CIFAR100   & 60000      & 32x32x3     & 100  \\
DoubleMoon & 5000       & 2           & 2 \\
EuroSAT & 27000 & 32x32x3 & 10 \\
EyeMovements & 10936 & 27 & 3\\
FashionMNIST & 70000 & 28x28 & 10 \\
House & 22784 & 16 & 2 \\
Jannis & 83733 & 54 & 4 \\
MNIST      & 70000      & 28x28       & 10  \\
MagicTelescope & 19020 & 10 & 2  \\
MiniBooNE & 130064 & 8 & 2 \\
NCI1       & 4110       & 37          & 2   \\
POL        & 15000      & 48          & 2   \\
Phoneme & 5404 & 5 & 2 \\
MiniBooNE  & 130064     & 50          & 2   \\
REDDIT-B   & 2000       & 1           & 2    \\
Spiral     & 5000       & 2           & 2  \\
SpiralHard & 10000       & 2           & 2  \\
\bottomrule
\end{tabular}
\end{table}

Hyper-parameter tuning was conducted in a structured and comparable manner across all baselines, while allowing for dataset-specific adaptations on the synthetic benchmarks. For all models and datasets, we fixed a common training protocol in terms of batch size 128, number of epochs 1000, optimizer AdamW with learning rate 0.01 and no weight decay. Model capacity was explored through grid searches over the hidden size $\{2,5,16\}$ and the number of layers $\{1,2\}$ (extended to $\{1,2,5\}$ for graph datasets), ensuring fairness across KAN, \method{}, and MLP baselines. On the synthetic datasets (DoubleMoon, Spiral, and SpiralHard), additional hyper-parameters were tuned to capture increasing geometric complexity. For KAN and \method{} models, we varied the number of fixed and starting number bases (KAN: $\{2,8,32,128\}$, iKAN: $\{2,64\}$), while for \method{} we further explored different bases weighting functions (Symmetric, OneSided, Truncated ERFC, and Truncated Exponential) and activation functions (ReLU, Sigmoid, PReLU, Tanh, GELU, SiLU). These choices were shared across the three synthetic datasets, and we fixed the Truncated Exponential and PReLU for the remaining datasets as they clearly gave more stable and higher performances. By evaluating all approaches on identical datasets with consistent hyperparameter tuning budgets and evaluation protocols, we provide a comprehensive assessment of whether adaptive basis learning offers advantages over both fixed-basis approaches and traditional neural networks across varying problem complexities.

\section{Ablation studies}
\label{sec:ablation}

\begin{figure*}[t]
\centering
\includegraphics[width=0.9\linewidth]{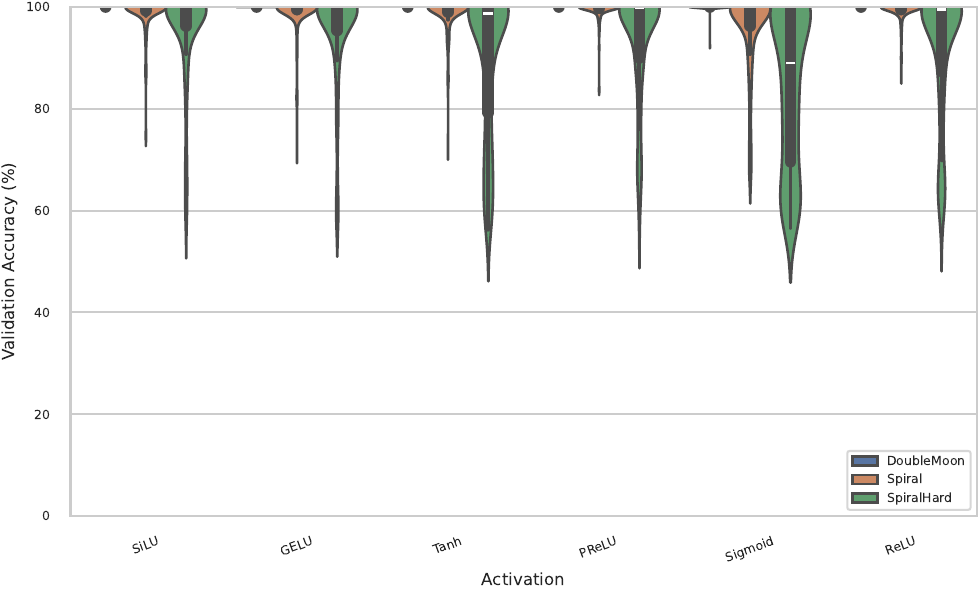}
\caption{Validation accuracy of \method{} across different activation functions (SiLU, GELU, Tanh, PReLU, Sigmoid, ReLU) on the synthetic datasets DoubleMoon, Spiral, and SpiralHard. Violin plots show the distribution of scores over multiple runs. All activation functions achieve near-perfect accuracy on DoubleMoon and Spiral, while SpiralHard exhibits greater variability, with Sigmoid showing the lowest and most variable performance.}
\label{fig:activation_toys}
\end{figure*}

\begin{figure*}[t]
\centering
\includegraphics[width=0.9\linewidth]{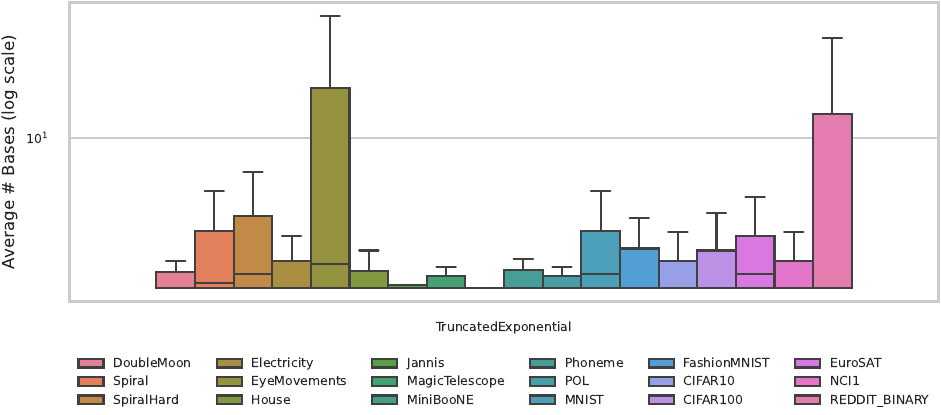}
\caption{Average number of basis functions learned by \method{} using the Truncated Exponential weighting distribution across all 18 datasets. The y-axis is shown on a log scale. Error bars indicate standard deviation over 10 runs. Most datasets converge to a small number of bases (below 10), while EyeMovements and REDDIT\_BINARY exhibit substantially higher and more variable basis counts, reflecting their greater complexity.}
\label{fig:basesByDistribution}
\end{figure*}

\begin{figure*}[t]
\centering
\includegraphics[width=0.9\linewidth]{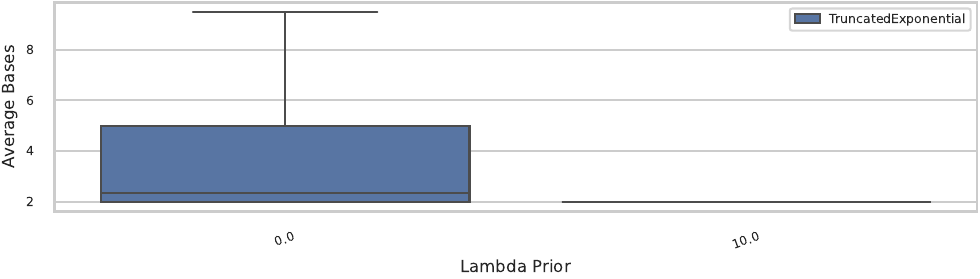}
\caption{Effect of the $\lambda$ prior hyperparameter on the average number of basis functions learned by \method{} using the Truncated Exponential weighting function. With $\lambda = 0.0$, the model exhibits high variability in the learned number of bases, while a stronger prior ($\lambda = 10.0$) regularizes the model toward fewer bases with reduced variance.}
\label{fig:lambda_prior_bases}
\end{figure*}

\begin{figure*}[t]
\centering
\includegraphics[width=0.9\linewidth]{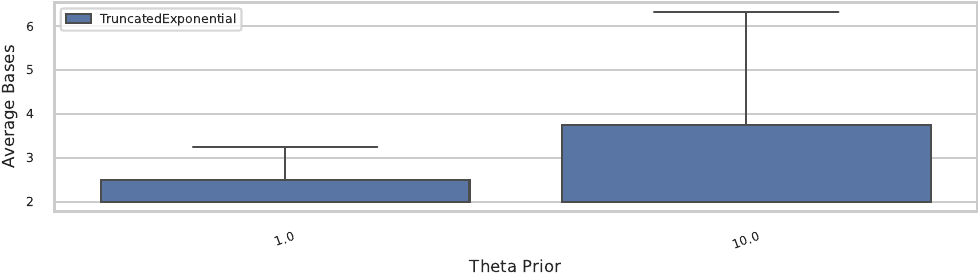}
\caption{Effect of the $\theta$ prior hyperparameter on the average number of basis functions learned by \method{} using the Truncated Exponential weighting function. A weaker prior ($\theta = 1.0$) results in fewer bases with lower variability, while a stronger prior ($\theta = 10.0$) leads to a higher average number of bases with increased variance across runs.}
\label{fig:theta_prior_bases}
\end{figure*}

We conduct a series of ablation studies to understand the sensitivity of \method{} to key design choices, including the choice of activation function, the weighting distribution, and the prior hyperparameters $\lambda$ and $\theta$.

\paragraph{Activation functions.}
We first investigate the impact of different activation functions on the performance of \method{}. \cref{fig:activation_toys} presents the validation accuracy across six activation functions (SiLU, GELU, Tanh, PReLU, Sigmoid, and ReLU) on the synthetic datasets. For the simpler DoubleMoon and Spiral tasks, all activation functions achieve near-perfect accuracy with minimal variance, indicating that the choice of activation is not critical for these problems. However, on the more challenging SpiralHard dataset, we observe greater differentiation: while most activations maintain high accuracy, Sigmoid exhibits notably lower and more variable performance. This suggests that for complex decision boundaries, smooth unbounded activations (such as PReLU and ReLU variants) provide more stable learning dynamics. Based on these findings, we selected PReLU as the default activation for subsequent experiments.

\paragraph{Learned number of bases across datasets.}
\cref{fig:basesByDistribution} shows the average number of basis functions learned by \method{} using the Truncated Exponential weighting distribution across all 18 datasets. The results reveal that most datasets converge to a compact representation with fewer than 10 basis functions, demonstrating that \method{} effectively identifies parsimonious models. However, certain datasets--particularly EyeMovements and REDDIT\_BINARY--exhibit substantially higher and more variable basis counts. This behavior reflects the inherent complexity of these tasks: EyeMovements involves time-series patterns with temporal dependencies, while REDDIT\_BINARY captures graph-structural information that may require richer functional representations. The variability in basis counts across runs for these datasets suggests that multiple model configurations can achieve similar performance, highlighting the exploratory nature of the variational optimization.

\paragraph{Effect of prior hyperparameters.}
We analyze the sensitivity of \method{} to the prior hyperparameters $\lambda$ and $\theta$, which control the regularization of the rate parameter and basis coefficients, respectively. \cref{fig:lambda_prior_bases} shows that the $\lambda$ prior has a strong regularizing effect on the learned number of bases. With $\lambda = 0.0$ (no regularization), the model exhibits high variability in the learned basis count, ranging from approximately 2 to 10 bases. In contrast, setting $\lambda = 10.0$ constrains the model to consistently learn around 2 bases with minimal variance. This demonstrates that the $\lambda$ prior provides an effective mechanism for controlling model complexity when prior knowledge about the desired sparsity is available.

The $\theta$ prior exhibits a different behavior, as shown in \cref{fig:theta_prior_bases}. A weaker prior ($\theta = 1.0$) results in fewer bases (median around 2.5) with relatively low variability. Increasing the prior strength to $\theta = 10.0$ leads to a higher average number of bases (median around 4) with substantially increased variance. This counterintuitive effect may arise because a stronger $\theta$ prior encourages larger coefficient magnitudes, which in turn requires more basis functions to adequately represent the target function. These ablation results provide practical guidance for practitioners: the $\lambda$ prior offers direct control over model sparsity, while the $\theta$ prior influences the trade-off between basis count and coefficient magnitude.

\end{document}